\documentclass[runningheads]{llncs}
\usepackage[T1]{fontenc}
\usepackage{graphicx}
\usepackage{amsmath,amssymb,amsfonts}
\usepackage{algorithmic}
\usepackage{textcomp}
\usepackage{multirow}
\usepackage{bbm}
\usepackage{booktabs}
\usepackage{color, colortbl}
\usepackage{floatrow}
\usepackage{bbding}
\newfloatcommand{capbtabbox}{table}[][\FBwidth]
\begin{document}
\title{Concept-Attention Whitening for Interpretable Skin Lesion Diagnosis}
%
%
\author{Junlin Hou\inst{1} \and
Jilan Xu\inst{2} \and
Hao Chen\inst{1,3}
}

\authorrunning{J. Hou et al.}
%
\institute{The Hong Kong University of Science and Technology, Hong Kong, China 
\and
Fudan University, Shanghai, China
\and
HKUST Shenzhen-Hong Kong Collaborative Innovation Research Institute, Shenzhen, China
}

\maketitle              
\begin{abstract}

The black-box nature of deep learning models has raised concerns about their interpretability for successful deployment in real-world clinical applications. 
To address the concerns, eXplainable Artificial Intelligence (XAI) aims to provide clear and understandable explanations of the decision-making process. 
In the medical domain, concepts such as attributes of lesions or abnormalities serve as key evidence for deriving diagnostic results.
Existing concept-based models mainly depend on concepts that appear independently and require fine-grained concept annotations such as bounding boxes. 
However, a medical image usually contains multiple concepts, and the fine-grained concept annotations are difficult to acquire. 
In this paper, we aim to interpret representations in deep neural networks by aligning the axes of the latent space with known concepts of interest.
We propose a novel Concept-Attention Whitening (CAW) framework for interpretable skin lesion diagnosis. 
CAW is comprised of a disease diagnosis branch and a concept alignment branch. In the former branch, we train a convolutional neural network (CNN) with an inserted CAW layer to perform skin lesion diagnosis. 
The CAW layer decorrelates features and aligns image features to conceptual meanings via an orthogonal matrix.
In the latter branch, the orthogonal matrix is calculated under the guidance of the concept attention mask.  
We particularly introduce a weakly-supervised concept mask generator that only leverages coarse concept labels for filtering local regions that are relevant to certain concepts, improving the optimization of the orthogonal matrix.
Extensive experiments on two public skin lesion diagnosis datasets demonstrated that CAW not only enhanced interpretability but also maintained a state-of-the-art diagnostic performance.

\keywords{Explainable AI  \and Concept Attention \and Skin Lesion Diagnosis}
\end{abstract}
%
%
%

\section{Introduction}
Deep learning has achieved significant advancements in medical image analysis. However, the black-box nature of deep learning greatly hinders its practical deployment and application \cite{kermany2018identifying,esteva2017dermatologist}. The networks usually output predictions without providing any explanation, resulting in a lack of interpretability. Therefore, there is an urgent need to develop eXplainable AI (XAI) techniques that can enhance the transparency and understandability of the decision-making process.

Recently, there has been an increasing consensus that XAI should incorporate explanations based on \textit{concepts} \cite{poeta2023concept}. In the medical domain, concepts can be defined as high-level attributes of lesions or abnormalities, serving as evidence for deriving diagnostic results. For instance, \textit{blue whitish veil}, \textit{atypical pigmentation network}, and \textit{irregular streaks} can be important concepts for diagnosing melanoma skin disease \cite{Kawahara2018-7pt}.
Given the concept annotations, current ante-hoc concept-based models mainly fall into two categories. The first is joint training for the target task and concept representation learning \cite{koh2020cbm,espinosa2022cem,kim2023probcbm}. 
For example, Concept Bottleneck Model (CBM) \cite{koh2020cbm} first predicted concepts and then made final predictions, but its generalization capability is much lower than standard end-to-end models. 
The second is to inject concepts from an external concept dataset to train deep neural networks by modifying a middle layer to represent concepts \cite{chen2020concept,yuksekgonul2022pcbm,rigotti2021ct,kazhdan2020cme}. 
Chen \emph{et al.} \cite{chen2020concept} proposed a concept whitening layer, consisting of whitening and orthogonal transformation to align the concepts with the axes.  
These methods achieve good interpretability on natural images due to two primary reasons. 
First, a majority of images contain only one concept that is highly related to the category (\emph{e.g.}, airplane$\rightarrow$airfield). 
Second, they have fine-grained concept annotations (\emph{e.g.}, bounding boxes) to crop concept regions from raw images. In this way, a set of most representative images that depict the concept can be collected to generate precise concept features.

However, medical images present a more complex challenge compared to natural images. 
A medical image usually contains multiple concepts, such as different types of lesions or abnormalities, that need to be considered for accurate classification. 
Moreover, obtaining fine-grained concept annotations such as masks or bounding boxes for medical images is time-consuming and laborious. The available concept information is often limited to coarse, image-level concept labels.
To address the above issues, we propose a novel method named Concept-Attention Whitening (CAW) to enhance the representation interpretability of skin lesion diagnosis, where the axes of latent space are aligned with specific concepts.
The main contributions of our method are three folds: 
(1) We establish a unique dual-branch optimization framework. In the disease diagnosis branch, given a disease dataset, we train a convolutional neural network (CNN) to perform disease classification with a novel CAW layer inserted. 
The CAW layer aims to decorrelate features by whitening transformation and assign conceptual meanings to specified dimensions via an orthogonal matrix.
(2) In the concept alignment branch, we use a concept dataset to calculate the orthogonal matrix based on concept features. 
As an image may contain multiple concepts, we particularly introduce a weakly-supervised concept mask generator to produce concept-attentive masks by only using the concept labels. The concept mask highlights the most relevant local regions regarding a certain concept. 
In this way, we can obtain representative concept features and calculate an accurate orthogonal matrix by solving an optimization problem. 
(3) Extensive experiments were conducted on two skin lesion diagnosis datasets with concept annotations. 
Compared with existing state-of-the-art methods, our proposed method not only enhanced interpretability but also maintained a high diagnostic performance.

\section{Methodology}
\begin{figure*}[t]
\begin{center}
   \includegraphics[width=\linewidth]{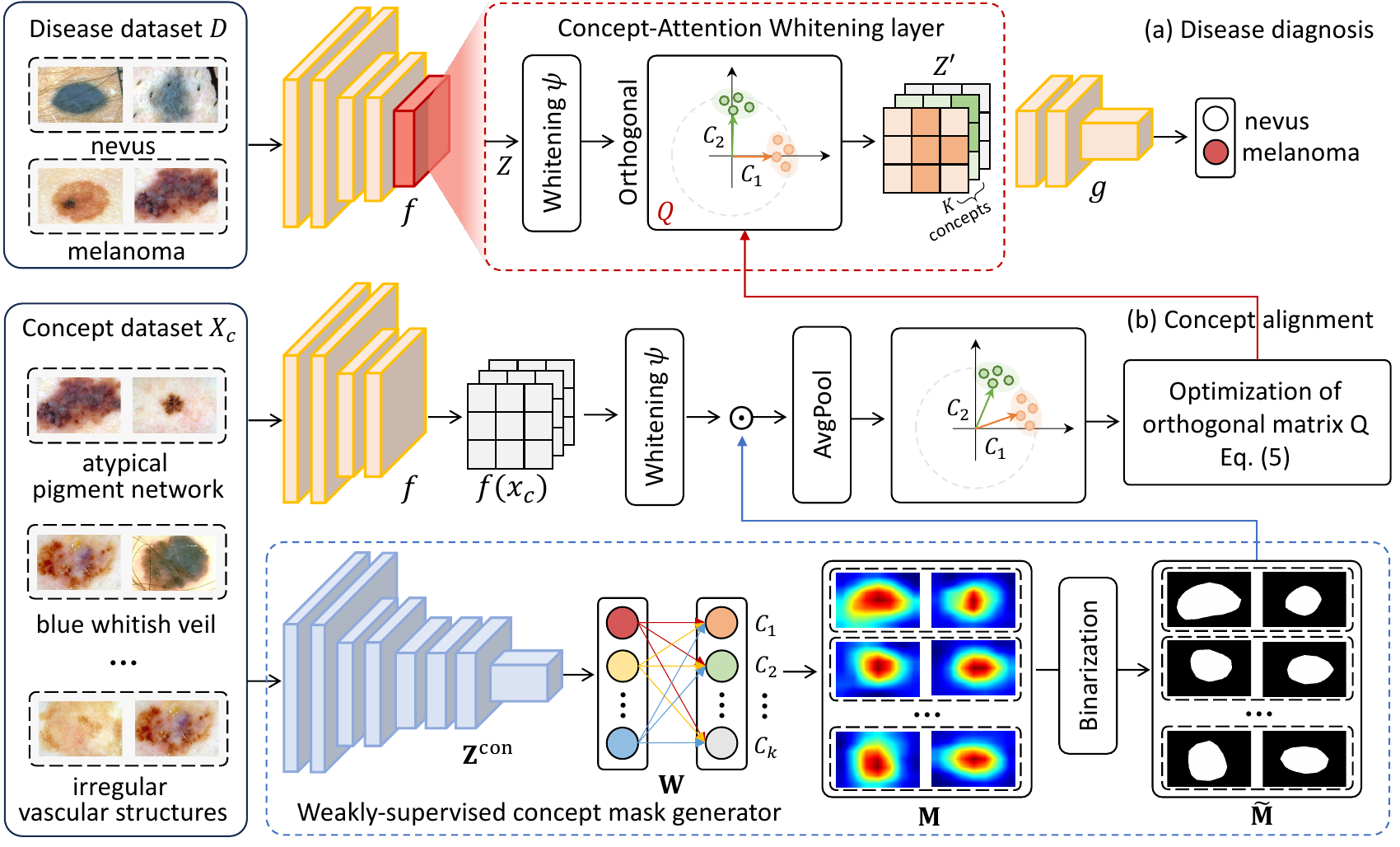}
\end{center}
\vspace{-1.0em}
   \caption{An overall framework of the proposed Concept-Attention Whitening (CAW), including (a) a disease diagnosis branch and (b) a concept alignment branch.}
\label{fig:framework}
\end{figure*}

Our overall objective is to train an interpretable disease diagnosis model that satisfies: (1) high disease diagnosis performance; and (2) the latent image features produced by the model are aligned to a pre-defined set of concepts. 
To achieve this goal, we propose a novel Concept-Attention Whitening (CAW) framework for interpretable skin lesion diagnosis using clinical concepts, as illustrated in Fig. \ref{fig:framework}.
Specifically, a Concept-Attention Whitening layer is inserted into the encoder network to disentangle concepts and align the latent image features with known concepts of interest. 
For network training, we adopt a dual-branch architecture, comprised of disease diagnosis and weakly-supervised concept alignment.

\subsection{Disease Diagnosis Branch}
Given the disease dataset $D=\{(x_i,y_i)\}_{i=1}^N$ where $y_i$ is the disease label of sample $x_i$, we train a CNN (\emph{e.g.}, ResNet-50) to classify the skin disease. 
We replace the Batch Normalization (BN) layer with our Concept-Attention Whitening (CAW) layer to produce intepretable representations.
Let $Z\in\mathbbm{R}^{b\times d\times h\times w}$ be the feature map before the CAW layer, where $b, d, h,$ and $w$ denote the batch size, dimension, height, and width. 
The CAW layer is composed of two operations: 
(1) a whitening transformation to separate different concepts in the latent space; and (2) an orthogonal transformation to align the axes of latent space with pre-defined concepts. 
Next, we will introduce each operation in detail. 

\noindent\textbf{Whitening transformation.} 
First, we flatten the feature map $Z\in\mathbbm{R}^{b\times d\times h\times w}$ into shape $d\times n$, where $n=b\times h\times w$. 
Then, a whitening transformation $\psi$ is adopted to decorrelate and standardize the feature $Z$ by:
\begin{equation}
    \psi(Z)=W(Z-\mu \mathbf{1}_{1\times n}),
\end{equation}
where $\mu$ is the mean of $n$ samples; $W\in\mathbbm{R}^{d\times d}$ is the whitening matrix which can be calculated by ZCA algorithm \cite{huang2019iterative}. 
After whitening, each dimension of the feature becomes mutually independent.

\noindent\textbf{Orthogonal transformation.} 
In this step, we align each separated dimension to a specific concept. This is achieved by leveraging an orthogonal matrix $Q\in\mathbbm{R}^{d\times d}$, in which the column $q_k$ is defined as the feature of concept $c_k$. 
For the calculation of the matrix $Q$, we particularly propose a weakly-supervised concept alignment, which will be elaborated in the next section. 
As the whitening matrix $W$ is rotation-free, $Q^T W$ is also a valid whitening matrix. Thus after CAW, we can obtain the interpretable feature $Z'=Q^T\psi(Z)$ and reshape it into its original size $b\times d\times h\times w$ for subsequent computation.

Finally, the feature map $Z'$ is fed into the rest of the network to predict the disease label by the following objective:
\begin{equation}
    \min \frac{1}{N}\sum_{i=1}^N \mathcal{L}_{ce}(g(Q^T \psi(f(x_i))),y_i),
\end{equation}
where $f$ and $g$ are layers before and after the CAW layer respectively. $\psi$ is a whitening transformation. $Q$ is the orthogonal matrix. $\mathcal{L}_{ce}$ is the cross-entropy loss for skin disease classification.

\subsection{Concept Alignment Branch}

The concept alignment branch aims to estimate an orthogonal matrix $Q$ for the disease diagnosis branch by leveraging the concept dataset $X_c$.
Specifically, the concept dataset $X_c=\{X_{c_k}\}_{k=1}^K$ consists of $K$ subsets of images, where each $X_{c_k}$ represents the images with concept $c_k$. 
We develop a weakly-supervised concept mask generator to identify concept-attentive features, which serve as guidance for refining the orthogonal matrix $Q$.

\noindent\textbf{Weakly-supervised concept mask generator.}
First, we pre-train a concept classification network on the concept dataset $X_c$ supervised by concept labels. 
We generate concept activation maps from the pre-trained concept classification network in a weakly-supervised manner. 
Formally, given a concept feature map $\mathbf{Z}^{\mathrm{con}}\in \mathbbm{R}^{d\times h\times w}$, the weights of the classifier $\mathbf{W}\in\mathbbm{R}^{d\times K}$ can be regarded as the prototypes of $K$ concepts. 
With regard to the predicted concept class $c_k$, we select the corresponding prototype $\mathbf{W}^{c_k}\in\mathbbm{R}^d$ and measure its similarity with each pixel of the feature map $\mathbf{Z}^{\mathrm{con}}$. The activation value of  $\mathbf{M}_{c_k}(i,j)$ at spatial location $(i,j)$ is calculated by summing the multiplication of $\mathbf{W}^{c_k}$ and $\mathbf{Z}^{\mathrm{con}}(i,j)$ across the channel dimension: 
\begin{equation}
    \mathbf{M}_{c_k}(i,j)=\sum_d \mathbf{W}_d^{c_k} \cdot \mathbf{Z}^{\mathrm{con}}_d(i,j).
\end{equation}
The activation map $\mathbf{M}_{c_k}\in\mathbbm{R}^{h\times w}$ is normalized to the interval $[0, 1]$. We further binarize the concept map by a pre-defined threshold $\gamma$ to generate the concept mask, which is used to filter the most discriminative concept features: 
\begin{equation}
    \mathbf{\tilde{M}}_{c_k}(i,j)=\begin{cases}
        1, & \textrm{if} ~ \mathbf{M}_{c_k}(i,j) > \gamma \\
        0, & \textrm{otherwise}        
    \end{cases}.
\end{equation}
The concept mask $\mathbf{\tilde{M}}_{c_k}\in\{0,1\}^{h\times w}$ highlights one specific concept of interest $c_k$ in the image, and it is subsequently used to guide the optimization of $Q$.

\noindent\textbf{Optimization of $\mathbf{Q}$.}
To align the $k$-th feature dimension with concept $c_k$, we need to find an orthogonal matrix $Q\in\mathbbm{R}^{d\times d}$ whose column $q_k$ corresponds to the $k$-th axis, by optimizing the concept alignment objective:
\begin{equation}
\begin{aligned}
    \max_{q_1,q_2,...,q_K} \sum_{k=1}^K \frac{1}{|X_{c_k}|} &\sum_{x_{c_k}\in X_{c_k}} q_k^T \text{AvgPool}(\mathbf{\tilde{M}}_{c_k} \odot \psi(f(x_{c_k})))\\
    & s.t ~ Q^T Q=I_d
\end{aligned},
\end{equation}
where $\text{AvgPool}$ denotes average pooling over the spatial dimension, resulting in a concept-attentive feature vector with shape $d\times 1$. 
This optimization problem is a linear programming problem with quadratic constraints (LPQC), which is generally NP-hard. 
Since directly solving the optimal solution is intractable, we optimize it by gradient methods on the Stiefel manifold~\cite{wen2013feasible}. 
At each step $t$, the orthogonal matrix $Q$ is updated by Cayley transform:
\begin{equation}
    Q^{(t+1)}=(I+\frac{\eta}{2}A)^{-1}(I-\frac{\eta}{2}A)Q^{(t)},
\end{equation}
where $\eta$ is the learning rate, $A=G(Q^{(t)})^T-Q^{(t)}G^T$ is a skew-symmetric matrix and $G$ is the gradient of the loss function.

\section{Experiments}

\subsection{Datasets and Implementation Details}


\textbf{Datasets.} Derm7pt \cite{Kawahara2018-7pt} consists of 1,011 dermoscopic images, annotated with disease and clinical concept labels. Following \cite{patricio2023coherent,bie2024mica}, we filter the dataset to obtain a subset of 827 images. The disease categories include nevus and melanoma, and the concept categories cover 12 elements from the 7-point checklist \cite{argenziano1998epiluminescence}.
SkinCon \cite{daneshjou2022skincon} includes 3,230 images from the Fitzpatrick 17k skin disease dataset \cite{groh2021evaluating} that are densely annotated with 48 clinical concepts. The disease categories comprise malignant, benign, and non-neoplastic. 
We select 22 concepts that have at least 50 images representing the concept. 
The dataset is partitioned into 70\%, 15\%, and 15\% for training, validation, and testing, respectively.
The specific concepts used in the two datasets are enumerated in the Supplemental Material.

\noindent\textbf{Implementation details.} 
For the Derm7pt \cite{Kawahara2018-7pt} and SkinCon \cite{daneshjou2022skincon} datasets, we employ the ImageNet-pretrained ResNet-18 and ResNet-50 models \cite{he2016resnet} as the backbones, respectively. We replace the BN with CAW layer in the 8/16-th layer for ResNet-18/50. All images are resized to 224$\times$224. Data augmentation includes random horizontal flip, random cropping, and rotation.
The network is trained for 100 epochs with a batch size of 64 and a learning rate of 2e-3.
We adopt AUC, ACC, and F1 score as the evaluation metrics.

\subsection{Experimental Results}

\noindent\textbf{Skin disease diagnosis.}
We compare the performance of skin disease diagnosis of our proposed CAW with other state-of-the-art methods, as shown in Table \ref{table:result}. To establish an upper bound, we first train a standard black-box ResNet model without the interpretability of concepts. 
In comparison, the group of CBM-based models \cite{sarkar2022framework,yuksekgonul2022pcbm,patricio2023coherent,bie2024mica} demonstrates a significant performance decrease for skin disease diagnosis, despite enhancing interpretability. 
Notably, the vanilla CW model \cite{chen2020concept} shows superior diagnosis performance, surpassing all the CBM-based methods across all the evaluation metrics. With concept attention, our proposed CAW significantly further improves the skin disease diagnosis performance on both datasets. CAW approaches the performance of the black-box model in terms of AUC, and even surpasses it on ACC and F1.

\begin{table*}[t]
\begin{center}
\caption{Disease diagnosis results of concept-based state-of-the-art methods. We report the results as mean$_{\textrm{std}}$ of three random runs.}
\label{table:result}
\vspace{-1.0em}
\begin{tabular}{lccccccccccccc}
\toprule
\multirow{2}{*}{Method} & \multicolumn{3}{c}{Derm7pt} & \multicolumn{3}{c}{SkinCon}\\
\cmidrule(lr){2-4}\cmidrule(lr){5-7}
& AUC & ACC & F1 & AUC & ACC & F1 \\
\midrule
ResNet \cite{he2016resnet} & 
~89.48$_{0.46}$~ & ~84.48$_{0.47}$~ & ~80.60$_{0.69}$~ & ~80.85$_{0.71}$~ & ~78.85$_{0.57}$~ & ~77.80$_{0.64}$~\\
\midrule
Sarkar et al. \cite{sarkar2022framework} & 76.22$_{2.06}$ & 73.89$_{1.47}$ & 66.81$_{1.23}$ & 68.21$_{1.44}$ & 71.14$_{1.21}$ & 71.32$_{1.38}$\\
PCBM \cite{yuksekgonul2022pcbm} & 72.96$_{2.19}$ & 76.98$_{1.39}$ & 71.04$_{1.15}$ & 68.94$_{1.59}$ & 71.04$_{1.13}$ & 70.47$_{0.75}$\\
PCBM-h \cite{yuksekgonul2022pcbm} & 83.27$_{1.14}$ & 79.89$_{0.89}$ & 74.48$_{1.37}$ & 69.53$_{1.67}$ & 72.28$_{1.39}$ & 72.28$_{1.29}$ \\
CBE \cite{patricio2023coherent} & 76.60$_{0.35}$ & 83.75$_{0.26}$ & 78.13$_{0.44}$ & 72.75$_{1.15}$ & 73.75$_{1.10}$ & 73.56$_{1.31}$\\
MICA w/ bot \cite{bie2024mica} & 84.11$_{1.10}$ & 82.20$_{1.31}$ & 78.08$_{1.22}$ & 75.89$_{1.11}$ & 74.29$_{1.09}$ & 74.74$_{1.21}$\\
MICA w/o bot \cite{bie2024mica} & 85.59$_{1.11}$ & 83.94$_{0.99}$ & 79.38$_{1.34}$ & 75.92$_{1.13}$ & 75.63$_{1.07}$ & 75.43$_{1.24}$ \\
\midrule
CW \cite{chen2020concept} & 86.50$_{0.40}$ & 83.85$_{0.48}$ & 80.00$_{0.75}$ & 79.49$_{0.60}$ & 78.28$_{0.57}$ & 77.30$_{0.67}$\\
CAW (ours) & \textbf{88.60}$_{0.10}$ & \textbf{84.79}$_{0.79}$ & \textbf{81.34}$_{0.85}$ & \textbf{80.47}$_{0.24}$ & \textbf{79.00}$_{0.19}$ & \textbf{77.76}$_{0.57}$\\
\bottomrule
\end{tabular}
\vspace{-2.0em}
\end{center}
\end{table*}

\begin{figure*}[t]
\begin{center}
   \includegraphics[width=\linewidth]{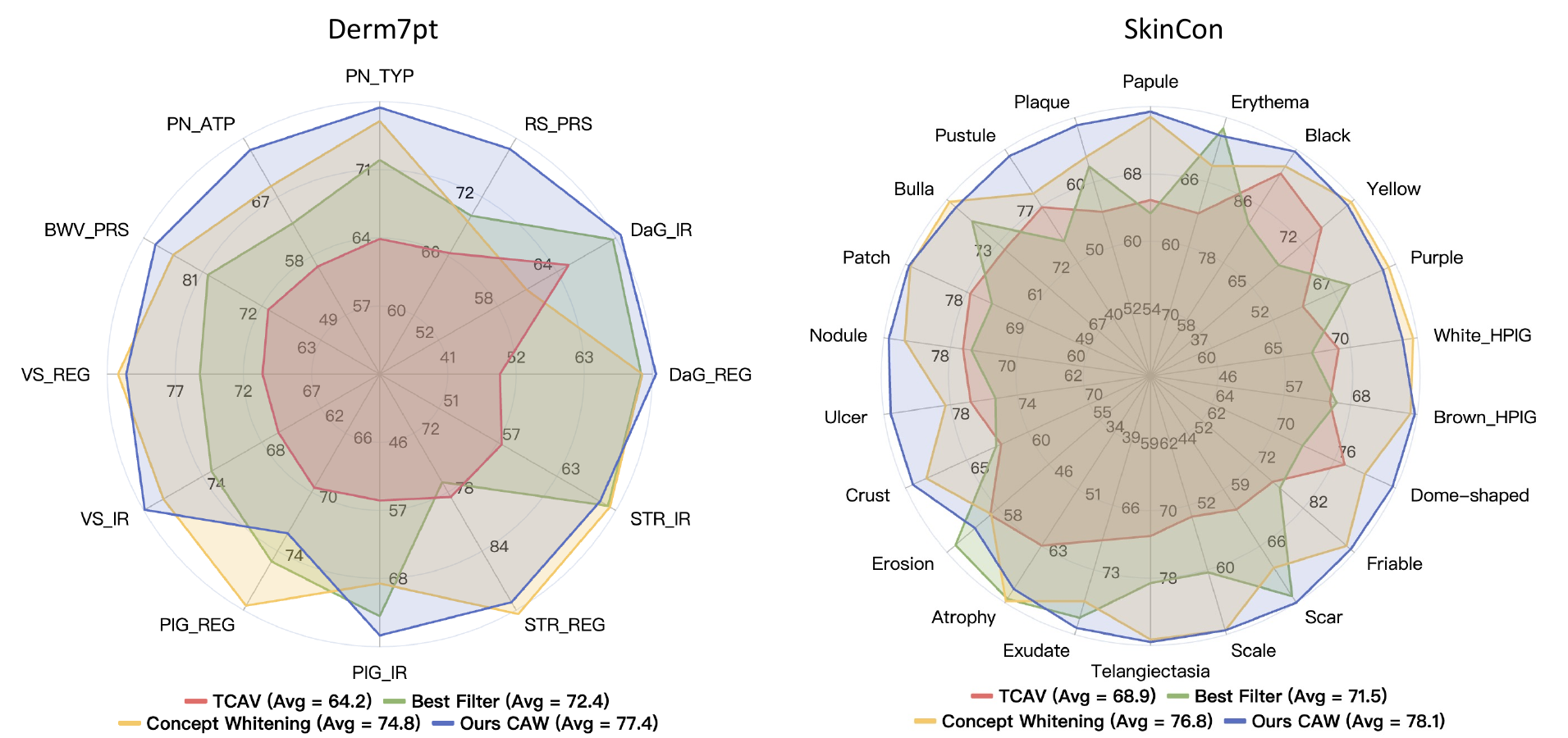}
\end{center}
\vspace{-1.0em}
   \caption{Comparison of concept detection on Derm7pt and SkinCon.}
\label{fig:concept detection}
\end{figure*}

\noindent\textbf{Concept detection.}
We conduct concept detection to measure the interpretability of our CAW quantitatively. Following \cite{chen2020concept}, we calculate the one-vs-all test AUC score of classifying the target concept in the latent space. 
We compare our CAW with the concept vectors learned by TCAV \cite{kim2018tcav} from black-box models, the filters in standard CNNs \cite{filter}, and CW \cite{chen2020concept}. 
As illustrated in Fig. \ref{fig:concept detection}, our CAW outperforms all the other methods on the average AUC score, reaching 77.4\% on Derm7pt and 78.1\% on SkinCon. Moreover, our CAW demonstrates superior or comparable detection performance on most concepts.

\subsection{Ablation Study and Discussion}

\noindent\textbf{Effect of concept attention mask.}
We conduct an ablation study to investigate the effect of our generated concept mask. 
As shown in the first row of Table \ref{tab:ablation}, the baseline approach, which merely uses the raw image as the concept map, yields the poorest performance, as it fails to capture precise concept features. 
The random and center-gaussian maps appear to enhance the performance, which can be attributed to their roles as a source of Cutout augmentation \cite{cutout}. 
To take a step further, we trained a skin lesion segmentation model \cite{ronneberger2015unet} to generate lesion masks, effectively eliminating redundant regions in the images and emphasizing the entire lesion area, leading to improved performance. However, there still exists a difference between the lesion area and the concept regions.
Finally, our generated concept masks by thresholding on concept maps accurately localize concept regions, resulting in the best performance in terms of both disease diagnosis and concept detection.

\begin{table*}
\begin{floatrow}
\capbtabbox{
\vspace{-1.0em}
\begin{tabular}{lcccc}
\toprule
Method & Disease Diag.~ & Concept Det.\\
\midrule
raw image & 86.96 & 74.79 \\
gaussian map & 87.16 & 74.92\\
random map & 87.78 & 75.93 \\
lesion mask & 87.94 & 75.55 \\
concept map & 88.13 & 76.23\\
concept mask & \textbf{88.60} &\textbf{77.40}\\
\bottomrule
\end{tabular}
\vspace{-2.0em}
}{
 \caption{Ablation study on concept mask. }
 \label{tab:ablation}
}
\capbtabbox{
\vspace{-1.0em}
\begin{tabular}{lcccc}
\toprule
$\gamma$ & Disease Diag.~ & Concept Det.\\
\midrule
0 & 86.95 & 74.79\\
0.2 & 87.68 & 77.12\\
0.5 & \underline{88.60} & \textbf{77.40}\\
0.6 & 88.44 & \underline{77.21}\\
0.8 & \textbf{88.63} & 76.52\\
1.0 & 87.92 & 76.03\\
\bottomrule
\end{tabular}
\vspace{-2.0em}
}{
 \caption{Analysis on threshold.}
 \label{tab:threshold}
 \small
}
\end{floatrow}
\end{table*}

\noindent\textbf{Analysis on threshold.}
We also investigate the impact of different binarization threshold values for concept mask generation. 
The results in Table \ref{tab:threshold} demonstrate that the performance consistently maintains a high level within the intermediate range of 0.5 to 0.8, indicating the robustness of our model to the choice of threshold. 
It is believed that a very small threshold would lead to the presence of concept-irrelevant regions, while a large threshold results in the loss of potential key concept information.

\begin{figure*}[t]
\begin{center}
   \includegraphics[width=\linewidth]{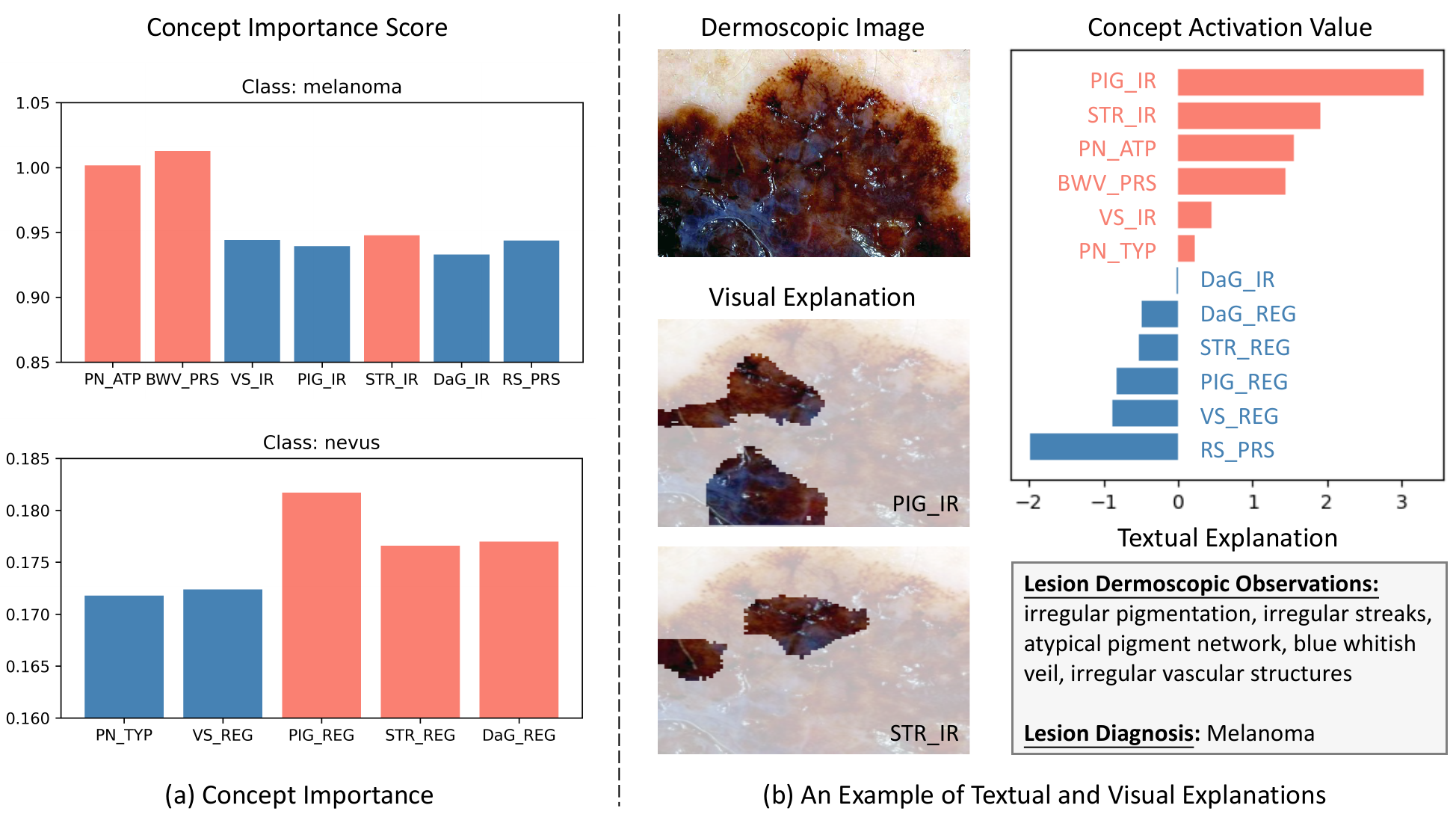}
\end{center}
\vspace{-1.0em}
   \caption{(a) Analysis on concept importance. (b) An example of explanations.}
\label{fig:concept importance}
\end{figure*}

\noindent\textbf{Concept importance.}
We measure the contribution of concepts to the disease diagnosis. 
The importance of the $k$-th concept $CI_k$ is defined as the ratio between the switched loss and the original loss \cite{fisher2019all}, which is given by $CI_k={loss}^{k}_{switch}/{loss}_{orig}$.
Here, ${loss}_{orig}$ denotes the original loss produced by the network without any permutation. 
To calculate the switched loss ${loss}^{k}_{switch}$, we randomly permute the $k$-th value of the concept feature along the batch dimension, \emph{i.e.}, replacing the $k$-th concept value of the current sample with another one from the batch. 
This indicates that the switched loss is expected to be large if the $k$-th concept is important for the current sample.
We show the concept importance scores on Derm7pt. 
The results depicted in Fig. \ref{fig:concept importance} (a) indicate the top three important concepts for melanoma and nevus, respectively, which confirm the consistency with the findings of dermatologists \cite{menzies1996sensitivity}.

\noindent\textbf{Example of explanation.}
We present an example that demonstrates how our CAW can offer comprehensible explanations during the disease diagnosis process, as illustrated in Fig. \ref{fig:concept importance} (b). For an input image, the activation value of its representation can be interpreted as the probability associated with a certain concept. Based on the activation values, a textual explanation can be derived to describe the image, and visual explanations can be generated simultaneously to emphasize the specific concept regions.

\section{Conclusion}
In this work, we propose an intrinsic interpretable XAI model based on concept, \emph{i.e.}, Concept-Attention Whitening (CAW), for skin lesion diagnosis. CAW consists of a disease diagnosis branch and a concept alignment branch. We specifically incorporate a weakly-supervised concept mask generator to filter the most relevant local regions, benefiting precise optimization of the orthogonal matrix in the CAW layer. Experiments on two skin lesion diagnosis datasets demonstrated the interpretability and superior diagnostic performance of CAW. 
In future work, we will consider the correlation of concepts by softening the orthogonality constraints in CAW, which is expected to promote the discovery of new concepts.

\begin{credits}
\subsubsection{\ackname} This work was supported by the HKUST (Project No. FS111) and Project of Hetao Shenzhen-Hong Kong Science and Technology Innovation Cooperation Zone (HZQB-KCZYB-2020083).

\subsubsection{\discintname}
The authors have no competing interests to declare that are relevant to the content of this article. 
\end{credits}
\bibliographystyle{splncs04}
\bibliography{mybibliography}




\end{document}


\title{Concept-Attention Whitening for Interpretable Skin Lesion Diagnosis}
%
%

\author{Supplementary Material
}

%
\institute{}
\maketitle              
%

%
%
%

\section{Concepts on Derm7pt and SkinCon datasets}




\begin{table*}
\begin{floatrow}
\capbtabbox{
\vspace{-1.0em}
\begin{tabular}{lll}
\toprule
& Concept & Abbreviation \\
\midrule
1 & typical pigment network & PN\_TYP \\
2 & atypical pigment network & PN\_ATP \\ 
3 & blue whitish veil & BWV\_PRS \\
4 & regular vascular structures & VS\_REG \\
5 & irregular vascular structures ~~ & VS\_IR \\
6 & regular pigmentation & PIG\_REG \\
7 & irregular pigmentation & PIG\_IR \\
8 & regular streaks & STR\_REG \\
9 & irregular streaks & STR\_IR \\
10 & regular dots and globules & DaG\_REG \\
11 & irregular dots and globules & DaG\_IR \\
12 & regression structures & RS\_PRS \\
\bottomrule
\end{tabular}
\vspace{-2.0em}
}{
 \caption{Concepts on Derm7pt.}
 \label{tab:ablation}
}
\capbtabbox{
\vspace{-1.0em}
\begin{tabular}{llll}
\toprule
& Concept & & Concept \\
\midrule
1 & Papule & 13 & Scale \\
2 & Plaque & 14 & Scar \\ 
3 & Pustule & 15 & Friable \\
4 & Bulla & 16 & Dome-shaped\\
5 & Patch & \multirow{2}{*}{17} & Brown-Hyper\\
6 & Nodule & & pigmentation\\
7 & Ulcer & \multirow{2}{*}{18} & White-Hypo\\
8 & Crust & & pigmentation\\
9 & Erosion & 19 & Purple\\
10 & Atrophy & 20 & Yellow\\
11 & Exudate & 21 & Black\\
12 & Telangiectasia & 22 & Erythema\\

\bottomrule
\end{tabular}
\vspace{-2.0em}
}{
 \caption{Concepts on SkinCon.}
 \label{tab:threshold}
 \small
}
\end{floatrow}
\end{table*}